\documentclass[letterpaper, 10 pt, conference]{ieeeconf}
\usepackage{amsmath}
\usepackage{mathtools}
\usepackage{algorithm}
\usepackage[noend]{algpseudocode}
\usepackage{hyperref}
\usepackage{multirow}

\usepackage{color,soul}

\usepackage{subcaption}
\usepackage{kantlipsum} 
\usepackage{mwe} 
\usepackage[squaren, Gray, cdot]{SIunits}
\usepackage{hhline}

\DeclareCaptionLabelSeparator{periodspace}{.\quad}
\IEEEoverridecommandlockouts                              

\title{\LARGE \bf Multirotor Planning in Dynamic Environments \\ using Temporal Safe Corridors}

\author{Charbel Toumieh and Alain Lambert
\thanks{The two authors are with the Universit\'e Paris-Saclay, CNRS, Laboratoire Interdisciplinaire des Sciences du Numérique, 91405, Orsay, France \url{https://www.lisn.upsaclay.fr}}%
}

\begin{document}

\maketitle
\thispagestyle{empty}
\pagestyle{empty}

\begin{abstract}
In this paper, we propose a new method for multirotor planning in dynamic environments. The environment is represented as a temporal occupancy grid which gives the current as well as the future/predicted state of all the obstacles. The method builds on previous works in Safe Corridor generation and multirotor planning to avoid moving and static obstacles. It first generates a global path to the goal that doesn't take into account the dynamic aspect of the environment. We then use temporal Safe Corridors to generate safe spaces that the robot can be in at discrete instants in the future. Finally we use the temporal Safe Corridors in an optimization formulation that accounts for the multirotor dynamics as well as all the obstacles to generate the trajectory that will be executed by the multirotor's controller. We show the performance of our method in simulations.
\end{abstract}

\section{INTRODUCTION}
\subsection{Problem statement}
Multirotor planning in dynamics environments has many real world industrial, humanitarian and military applications. That's why recent research efforts have been focused on providing solutions or solution elements to this challenging problem. 

A static environment has been the main assumption of multiple multirotor planning methods \cite{Bircheretal}, \cite{Bircheretal2}, \cite{burri2015real}, \cite{tordesillas2020faster}, \cite{gao2019flying}, \cite{gao2018optimal}, \cite{toumieh2020planning}, \cite{liu2017planning}, \cite{mellinger2011minimum} \cite{toumieh2022mace} \cite{toumieh2022near}.
It is the goal of this paper to present a new multirotor planner for dynamic environments.

\subsection{Related work} \label{sect:related_works}
Some works in the literature have addressed the problem of multirotor planning in dynamic environments using different approaches. Some consider only cooperative dynamic agents i.e. multi-agent planning, while assuming that the rest of the environment is static \cite{park2020efficient}, \cite{alonso2015collision}, \cite{kamel2017robust}, \cite{zhu2019chance}, \cite{Zhu2019bvc}, \cite{toumieh2021multiagent}. We will only discuss the works done where the dynamic obstacles are considered non-cooperative since this is the case that we treat in this paper.

In \cite{liu2018towards}, the authors propose a search based method to avoid collision with all kind of obstacles (dynamic obstacles, planning agents and static obstacles). However, search based methods in general result in a high computation time when planning complex maneuvers due to the curse of dimensionality. This renders them non suitable for real-time embedded applications.

In \cite{lin2020robust}, the authors model all obstacles (static and dynamic) as ellipses and include them in a non convex model predictive control formulation that keeps the planned discrete points outside the obstacle ellipses. While this approach may be suitable for a specific type of dynamic obstacles, decomposing the whole environment (including static obstacles) into ellipses is not trivial. Furthermore, decomposing a complex environment into ellipses might result in an inefficient representation or a large number of ellipses, which when added as constraints to the planner increase computation time. 

Finally, in \cite{tordesillas2020mader} the authors present a method for avoiding all types of obstacles (static and dynamic). They use a combination of an optimization method and a search-based method, where the output of the search-based method is given as an initial guess for the optimization method. This choice is made due to the fact that the optimization problem is non-convex and requires a good initial guess. This method represents obstacles as polyhedron. This representation is not trivial to generate from sensor measurements (camera images or lidar pointclouds).

\begin{figure}
\centering
\includegraphics[trim={2cm 0 0 0},clip,width=1\linewidth]{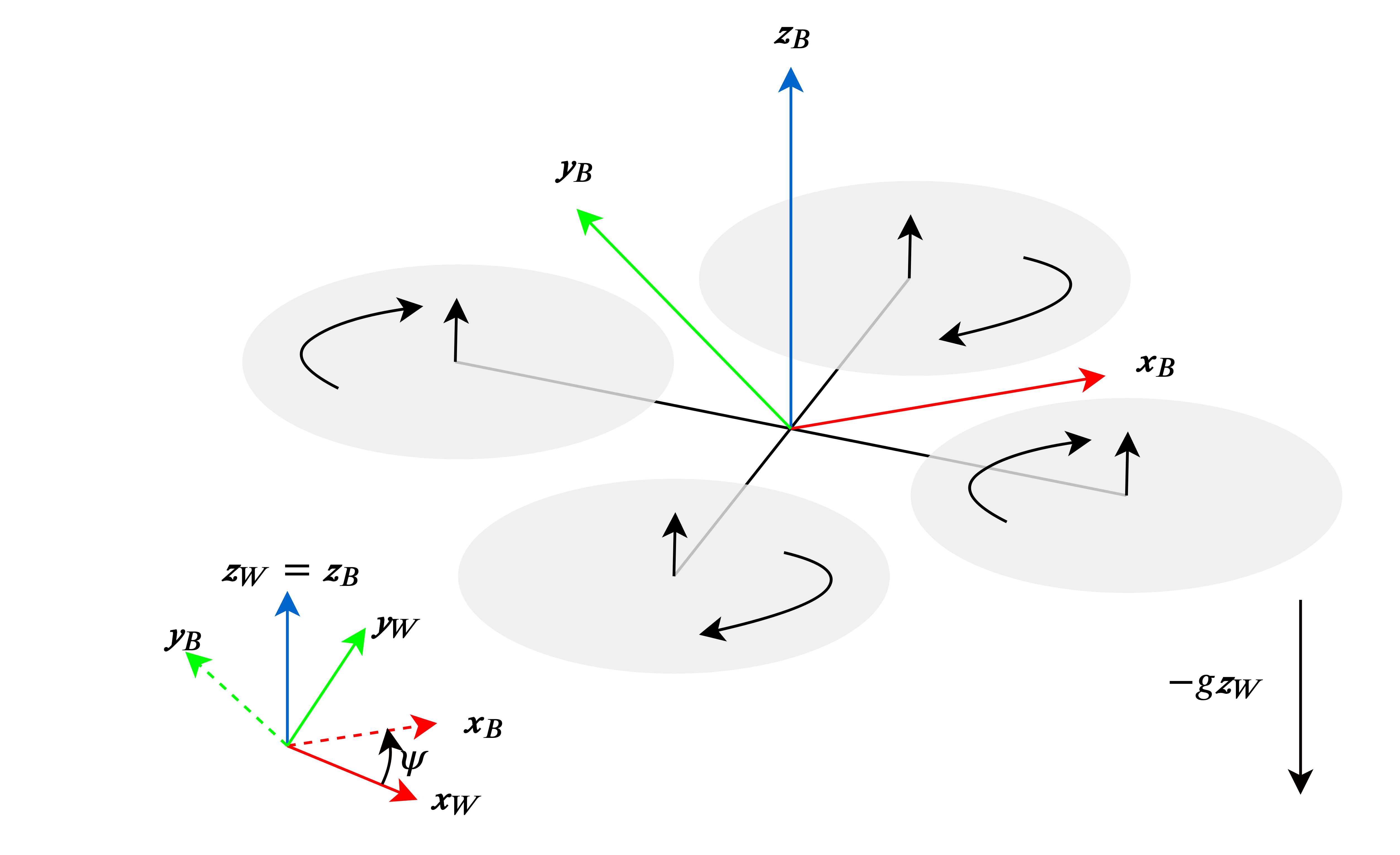}
\caption{Schematics of the multirotor model with the used
coordinate systems (body frame $B$ and world frame $W$).}
\label{fig:quad_fig}
\end{figure}

\subsection{Contribution}
The main contribution of this paper is a new planning framework for dynamic environments. The method takes as input a temporal occupancy grid i.e. a series of occupancy grids that represent all obstacle positions at discrete points in finite future time. We use the temporal occupancy grid to generate Temporal Safe Corridors which we then use to generate a trajectory that avoids all obstacles in a Model Predictive Control (MPC) fashion.

\section{Agent Model} \label{sect:model}

\begin{table}[h]
\caption{Nomenclature}
\label{tab:nomenclature}
\begin{center}
\begin{tabular}{c l}
\hline
$g$ & gravity \\
$m$ & multirotor mass \\
$\boldsymbol{p}$ & position vector $x,y,z$ in the world frame\\
$\boldsymbol{v}$ & velocity vector $v_x,v_y,v_z$ in the world frame\\
$\boldsymbol{a}$ & acceleration vector from thrust and gravity in the world frame\\
$\boldsymbol{j}$ & jerk vector $j_x,j_y,j_z$ in the world frame\\
$\boldsymbol{z}_W$ & world frame $z$\\
$\boldsymbol{z}_B$ & body frame $z$\\
$\boldsymbol{R}$ & rotation matrix from body to world frame\\
$\boldsymbol{D}$ & quadratic drag matrix\\
$\phi$ & roll angle\\
$\theta$ & pitch angle\\
$\psi$ & yaw angle\\
$c_{cmd}$ & total thrust command\\
\hline
\end{tabular}
\end{center}
\end{table}

Each multirotor is modeled as an agent whose control inputs are the attitude and thrust (the control input has the subscript $cmd$ in the equations of motion). The forces acting on the multirotor are the gravity, the drag forces and the thrust of the rotors.
The equations of motion are the following \cite{toumieh2020planning} (nomenclature Tab. \ref{tab:nomenclature}; Fig. \ref{fig:quad_fig}):
\begin{align}
     & \Dot{\boldsymbol{p}} = \boldsymbol{v} \\
     & \Dot{\boldsymbol{v}} = -g \boldsymbol{z}_W + \dfrac{c_{cmd}}{m} \boldsymbol{z}_B - \boldsymbol{R}\boldsymbol{D}\boldsymbol{R'}\boldsymbol{v} ||\boldsymbol{v}||_2 \label{eqn:acc}\\
     & \Dot{\phi} = \Dot{\phi}_{cmd} \\
     & \Dot{\theta} = \Dot{\theta}_{cmd} \\
     & \Dot{\psi} = \Dot{\psi}_{cmd} 
\end{align}

The above-mentioned equations are simplified 
into the following  linear model with the jerk $j$ as input to the system as in \cite{toumieh2020planning}:

\begin{equation}
  \label{eqn:mot}
    \begin{aligned}
     & \Dot{\boldsymbol{p}} = \boldsymbol{v} \\
     & \Dot{\boldsymbol{v}} = \boldsymbol{a} - \boldsymbol{D}_{lin\_max}\boldsymbol{v} \\
     & \Dot{\boldsymbol{a}} = \boldsymbol{j}
    \end{aligned}
\end{equation}

$\boldsymbol{D}_{lin\_max}$ is a diagonal matrix that represents the maximum linear drag coefficient in all directions. This matrix is identified offline as shown in \cite{toumieh2020planning}. The quadratic drag model is replaced with a linear worst-case scenario model. Many off-the-shelf solvers are very efficient for linear constraints, which is why we transformed the model into a linear one.

\section{The planner} \label{sect:the_method}
The planner take as input a temporal occupancy grid and then generates a trajectory that avoids all obstacles (static and dynamic) within the time horizon of the temporal occupancy grid. The planner is divided into 3 steps:
\begin{enumerate}
    \item Generating the Temporal Safe Corridor (TSC): in this step we take the temporal occupancy grid and use it to generate a TSC.
    \item Generating the reference trajectory: in this step we generate a global path from the position of the agent to the goal using the occupancy grid that represents all obstacles at the current instant. Then we use the path to generate a reference trajectory for the next step.
    \item Generating the safe trajectory: in this step we use the TSC and the reference trajectory in a Model Predictive Control (MPC)/ Mixed-Integer Quadratic Program (MIQP) formulation to generate a safe and locally optimal trajectory.
\end{enumerate}

\subsection{Generating the Temporal Safe Corridor} \label{sect:tsc}
\subsubsection{Temporal Occupancy Grids}
\begin{figure*}
\begin{subfigure}{1\textwidth}
\centering
\includegraphics[trim={0cm 0cm 0cm 0cm},clip,width=0.7\linewidth]{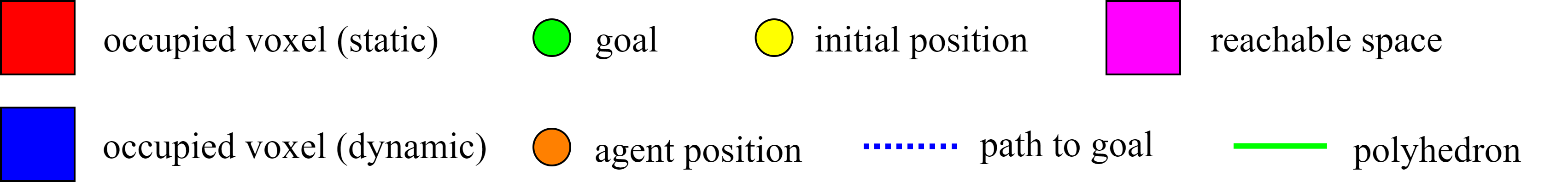}
\caption{Legend}
\label{fig:tog_legend}
\end{subfigure}

\begin{subfigure}{0.5\textwidth}
\centering
\includegraphics[trim={0cm 0cm 0cm -0.5cm},clip,width=0.8\linewidth]{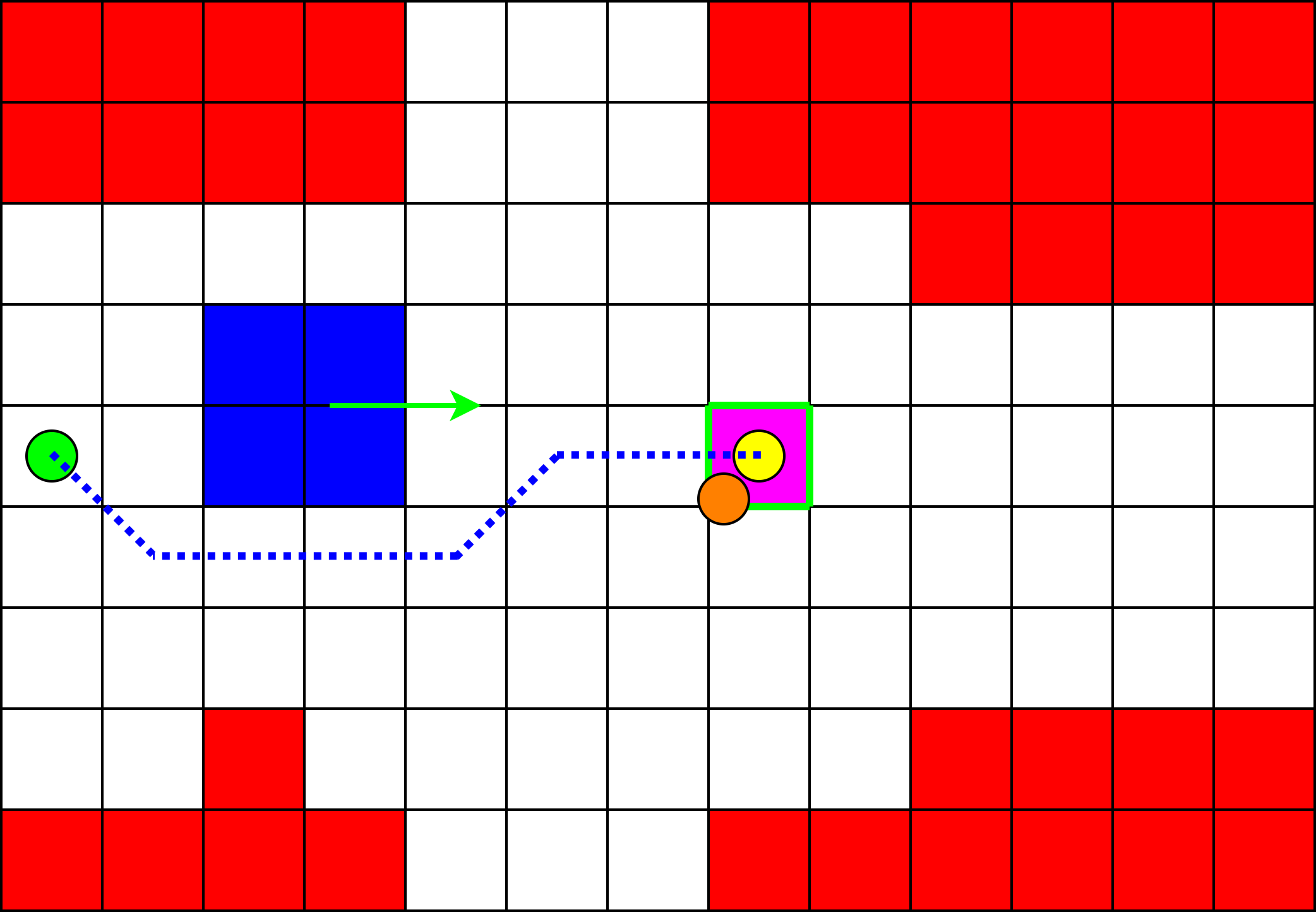}
\caption{$t = 0s$ to $t = 0.5s$}
\label{fig:tog_1}
\end{subfigure}
\begin{subfigure}{0.5\textwidth}
\centering
\includegraphics[trim={0cm 0cm 0cm -0.5cm},clip,width=0.8\linewidth]{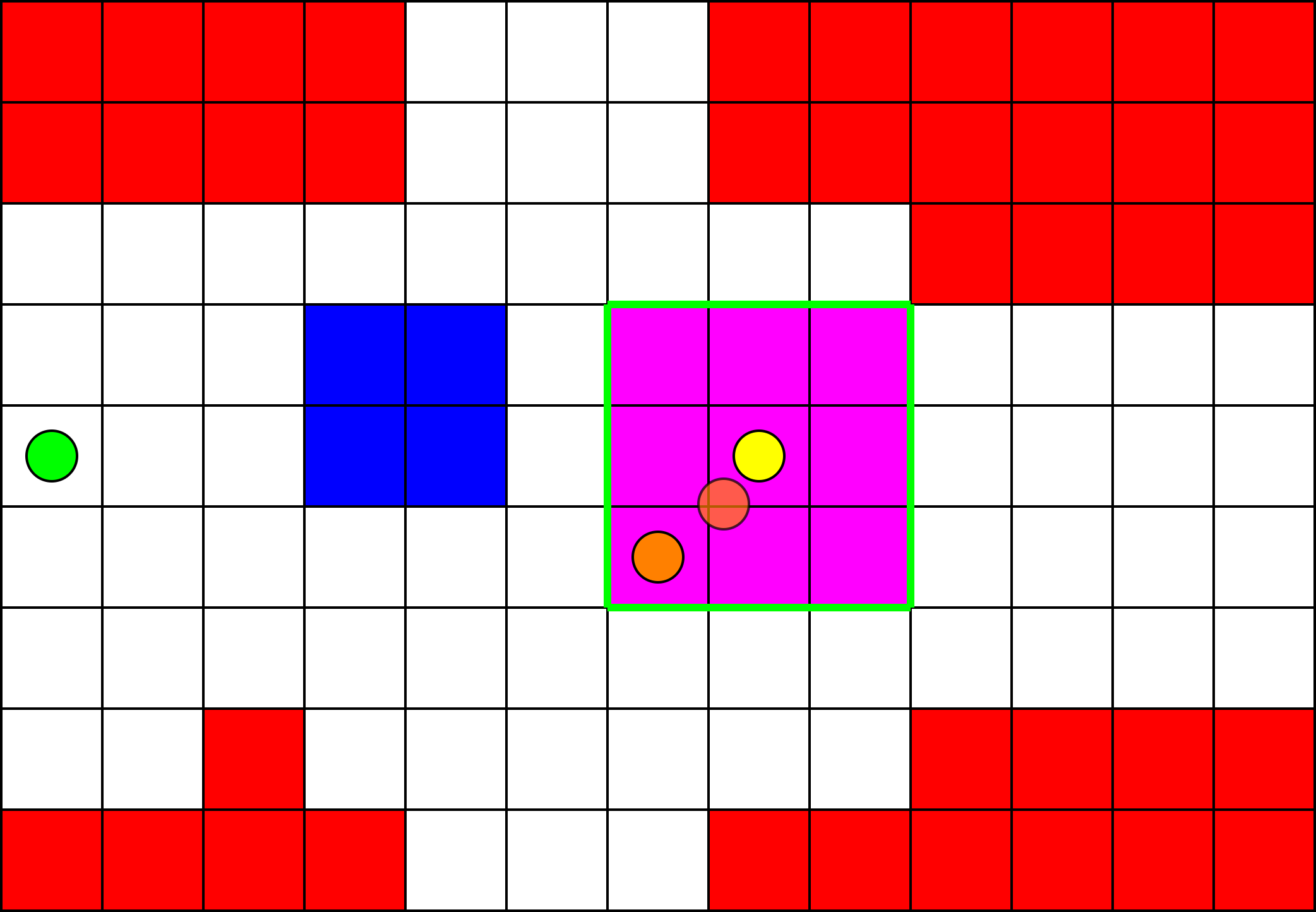}
\caption{$t = 0.5s$ to $t = 1s$}
\label{fig:tog_2}
\end{subfigure}

\begin{subfigure}{0.5\textwidth}
\centering
\includegraphics[trim={0cm 0cm 0cm -0.5cm},clip,width=0.8\linewidth]{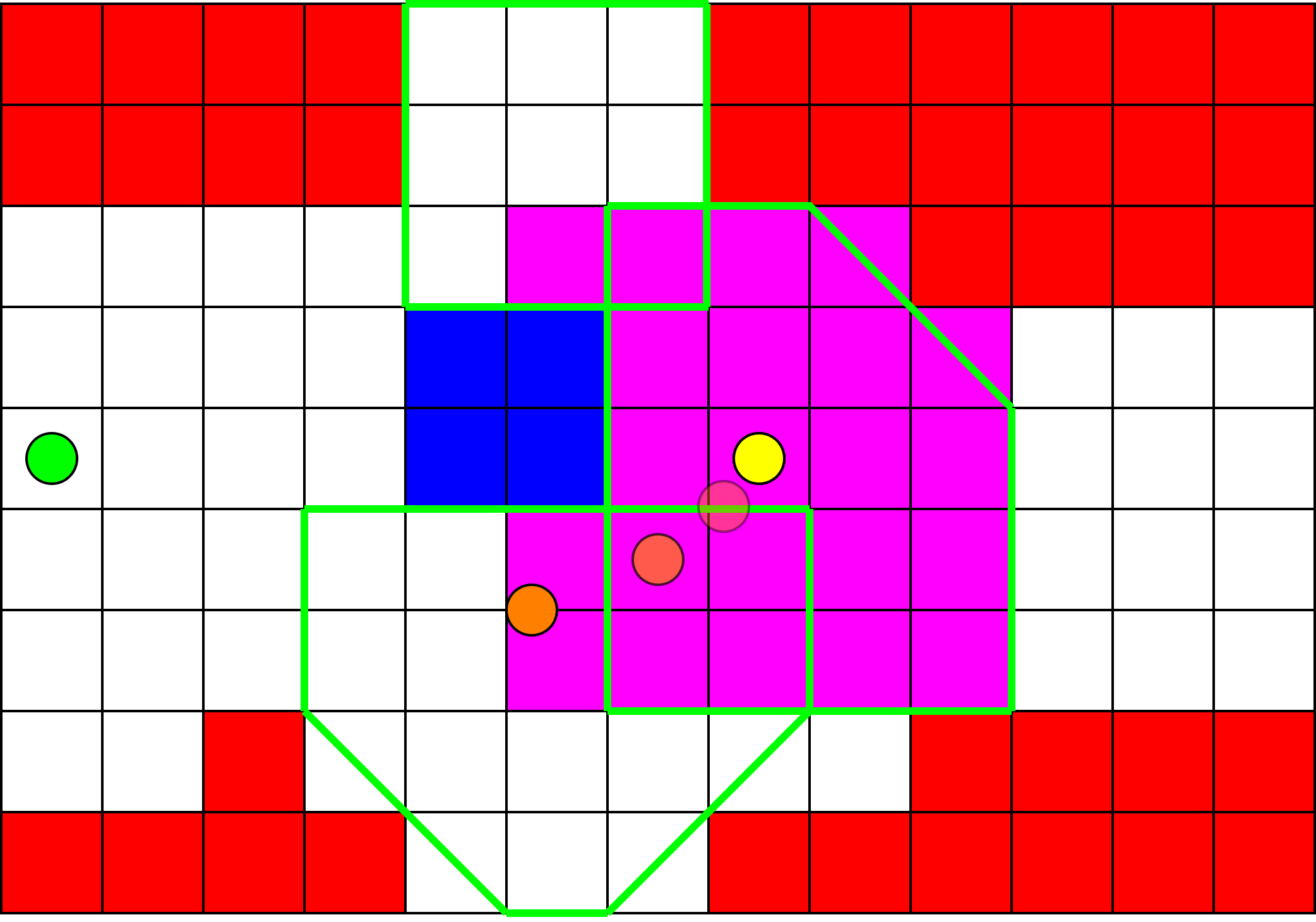}
\caption{$t = 1$ to $t = 1.5s$}
\label{fig:tog_3}
\end{subfigure}
\begin{subfigure}{0.5\textwidth}
\centering
\includegraphics[trim={0cm 0cm 0cm -0.5cm},clip,width=0.8\linewidth]{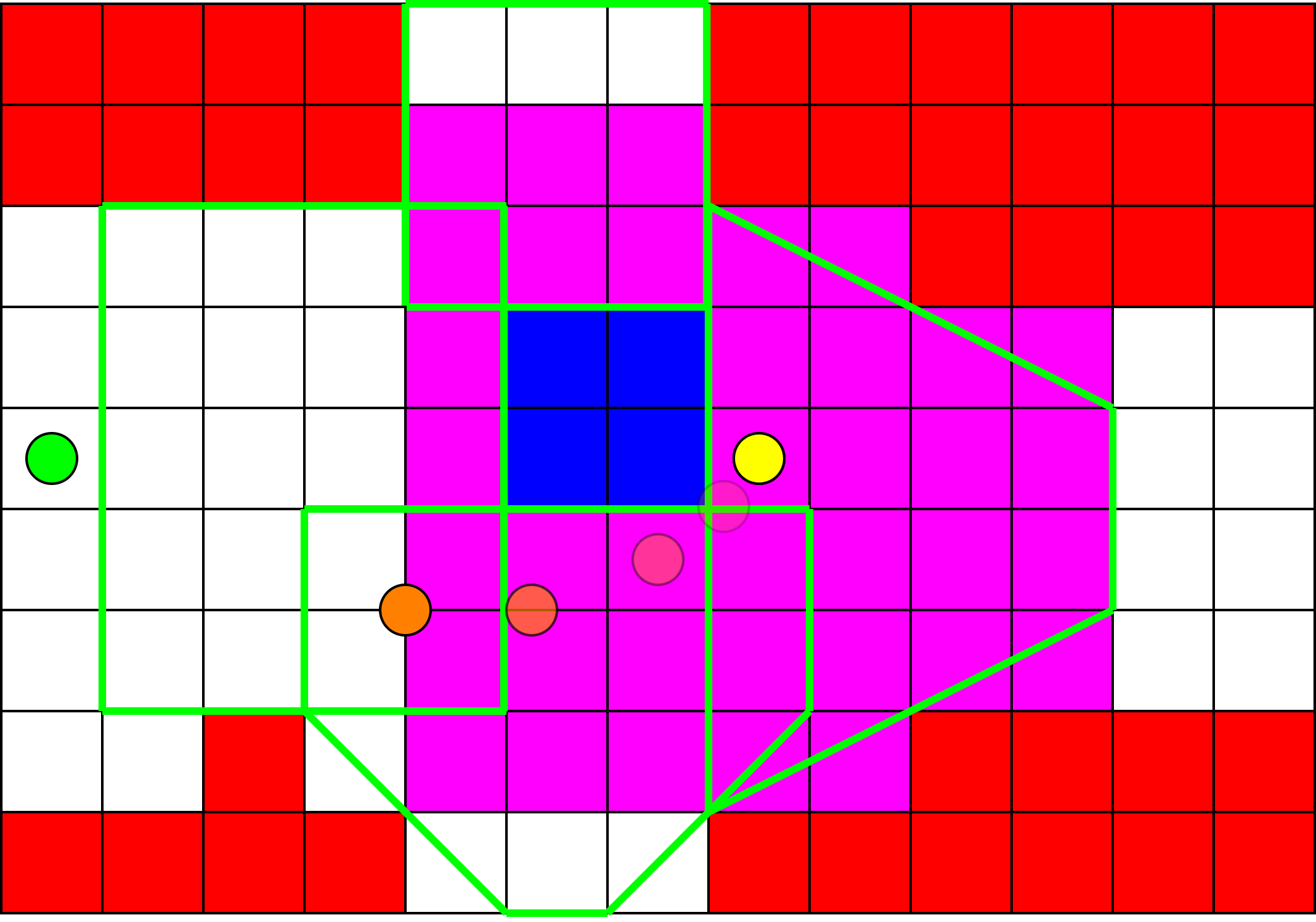}
\caption{$t = 1.5s$ to $t = 2s$}
\label{fig:tog_4}
\end{subfigure}

\caption{We show an example of a Temporal Occupancy Grid as well as a Temporal Safe Corridor over a duration of 2 seconds and a time step of $0.5s$. This means that the TOG is composed of 4 occupancy grids and the TSC is composed of 4 Safe Corridors. Each occupancy grid represents the occupied voxels during the whole duration of $0.5s$ i.e. all the position that the obstacle occupy during that duration. We show the occupied static obstacles in \textbf{red} and the occupied dynamic voxels in \textbf{blue}. The direction of movement of the obstacle is shown with a green arrow in Fig. \ref{fig:tog_1}. The initial position of the agent is shown as a \textbf{yellow} circle, and the current position is shown in \textbf{orange} (with previous positions being more transparent to indicate a position in the past). The goal that we want to reach is shown in \textbf{green}. The reachable space is shown in \textbf{magenta}. The polyhedra composing the Safe Corridors are shown in \textbf{green}. They are generated using the method described in \cite{toumieh2022shapeaware} which is built on \cite{toumieh2020convex}. The path generated in Sect. \ref{sect:ref_path} is shown in \textbf{dotted blue}. At each time step, the Safe corridor is generated to cover the reachable space. When planning, the position of the agent at each time step is constrained to be inside one of the polyhedra of the corresponding Safe Corridor.}
\label{fig:tog}
\end{figure*}

An occupancy grid partitions the space into regular cubes (voxels or cells) that are either occupied or free for the agent to move through. temporal occupancy grids (TOG)
are a series of occupancy grids that represents the state of the environment at discrete points in time for a given time horizon. Obstacles moving through time are captured by the change in the voxels that are occupied/free as shown in Fig. \ref{fig:tog}. This representation of the environment is showing promise especially in the autonomous driving domain \cite{hoermann2018dynamic}, \cite{jain2020discrete},  \cite{schreiber2020motion}.

\subsubsection{Temporal Safe Corridor}
For each occupancy grid of the TOG, we generate a series of polyhedra that covers free space that the agent can be in at that instant in time. The free space that we cover is the intersection between the reachable space and the free space in the occupancy grid. The reachable space depends on the agent dynamics and grows bigger as we step forward in time as shown in Fig. \ref{fig:tog}. The reachable space is generated using a simplified approach: we use the maximum velocity that the agent can have an multiply it by the time step ($0.5\ s$ in the example shown in Fig. \ref{fig:tog}) to obtain the maximum distance (which we denote $d_{reach}$) that the agent can traverse in that time duration. The reachable space in voxels corresponds to the voxels that cover $d_{reach}$ in all directions (a square with side length $d_{reach}$).

\subsection{Generating the reference trajectory} \label{sect:ref_path}
After generating the TSC, we generate a reference path that goes from the position of the agent to the goal. We use the first occupancy grid in the TOG (which corresponds to Fig. \ref{fig:tog_1}) to generate the path using Jumping Point Search (JPS) \cite{harabor2011online} and Distance Map Planner (DMP) \cite{jps3d}. JPS finds the shortest path to the goal while DMP pushes the path from the obstacles when possible to create a clearance margin. Note that this path does not take into account the moving aspect of the dynamic obstacles and considers them as static.

The path is then used to generate a local reference trajectory for the MIQP/MPC. At each iteration we sample $N$ points from the global trajectory to be used as reference for the $N$ discrete positions of the MIQP/MPC. At every iteration $l$, these points are sampled using a starting point $\boldsymbol{x}^{l}_{0,ref}$. At the first iteration this starting point is the position of the planning agent. From the starting position, we move along the global path at the sampling speed $v_{samp}$ (user input) for a time duration $h$, where $h$ is the time step of the MIQP/MPC. The point at which we arrive is the second reference point $\boldsymbol{x}^{l}_{1,ref}$. We continue sampling in the same fashion until we reach $\boldsymbol{x}^{l}_{N,ref}$. At the subsequent iterations, we use the optimal trajectory generated by the last step (MIQP/MPC) of the previous iteration: we check if the final state is close enough from the previous final reference point (within $thresh\_dist$). If yes, the local reference trajectory is generated with the above-mentioned algorithm starting from $\boldsymbol{x}^{l}_{0,ref} = \boldsymbol{x}^{l-1}_{1,ref}$. If no, the local reference trajectory generated at the previous iteration is used for the current iteration i.e. $\boldsymbol{x}^{l}_{0,ref} = \boldsymbol{x}^{l-1}_{0,ref}$.

\subsection{Solving the MIQP/MPC problem} \label{sect:form}
In this step, a collision-free optimal trajectory is generated. 
It uses the local reference trajectory generated in Sect. \ref{sect:ref_path} and the Temporal Safe Corridor generated in Sect. \ref{sect:tsc}. The error between the agent position and the local reference trajectory as well as the norm of the jerk are minimized. The resulting optimal trajectory gets us closer to the last sampled point
(Fig. \ref{fig:tog}). 

At every iteration $l$, the initial state $\boldsymbol{x}^{l}_0$ of the MPC is set to the second state
$\boldsymbol{x}^{l-1}_1$ of the last generated trajectory (except for the first iteration where the initial state is the initial robot position). The terminal velocity $\boldsymbol{v}_N$ is set to $\boldsymbol{0}$.

In case the solver fails to find a solution at a given iteration or the computation time exceeds the time step $h$, we skip the iteration (the solution is discarded), and at the next iteration, we solve the MIQP/MPC with the initial state $\boldsymbol{x}_2$ instead of $\boldsymbol{x}_1$. In case this also fails, we keep offsetting the initial position (which may reach $\boldsymbol{x}_N$ in the worst case).

\subsubsection{Dynamics} \label{sect:dynamics}
With $\boldsymbol{x} = [\boldsymbol{p} \ \boldsymbol{v} \ \boldsymbol{a}]^T $, $\boldsymbol{u} = \boldsymbol{j}$, $f(\boldsymbol{x}(t),\boldsymbol{u}(t))$ defined by Eq. \ref{eqn:mot}, the model is discretized using Euler or Runge-Kutta 4\textsuperscript{th}
order to obtain the discrete dynamics $\boldsymbol{x}_{k+1} = f_d(\boldsymbol{x}_k,\boldsymbol{u}_k)$.
We choose the Euler method as it results in faster solving times. With a discretization step of $h$, the discretized dynamics become:
\begin{equation}
    \begin{aligned}
     & \boldsymbol{p}_{k+1} = \boldsymbol{p}_{k} + h\boldsymbol{v_k} \\
     & \boldsymbol{v}_{k+1} = \boldsymbol{v}_{k} + h( \boldsymbol{a}_{k} -  \boldsymbol{D}_{lin\_max}\boldsymbol{v_k}) \\
     & \boldsymbol{a}_{k+1}= {a}_{k} + h\boldsymbol{j_k} \\
     & \boldsymbol{x}_{k} = [\boldsymbol{p}_{k} \  \boldsymbol{v}_{k} \ \boldsymbol{a}_{k}]^T \\
     & \boldsymbol{u}_k = \boldsymbol{j}_k
    \end{aligned}
\end{equation}

\subsubsection{State bounds}
The agent velocity is limited by the drag forces. The maximum bounds on the acceleration and the jerk in each direction are determined by the dynamics of the agent. We define $a_{x,max}$ and $a_{y,max}$ as the maximum L1 norm of the acceleration in the directions $x$ and $y$. We define $a_{z,max}$ and $a_{z,min}$ as the maximum and minimum values respectively of the acceleration in the $z$ direction. These values are deduced directly from the maximum thrust that a multirotor can generate. 

Finally, we define $j_{x,max}$, $j_{y,max}$ and $j_{z,max}$ as the maximum L1 norm of the jerk in the directions $x$, $y$ and $z$ respectively. These values represent the limits on the rotational dynamics of a multirotor.

\subsubsection{Collision avoidance} This is achieved by forcing every discrete point $k$ to be in one of the polyhedra of the corresponding SC of the TSC ($SC_k$).
Let's assume we have $P_k$ polyhedra in $SC_k$. They are described by
$\{(\boldsymbol{A}_{kp}, \boldsymbol{c}_{kp})\}$, $p = 0 : P-1$.
The constraint that the discrete position $\boldsymbol{p}_k$ is in a polyhedron $p$ is described by $\boldsymbol{A}_{kp}.\boldsymbol{p}_k \leq \boldsymbol{c}_{kp}$. We introduce binary variables $b_{kp}$ ($P_k$ variables for each $\boldsymbol{x}_k$, $k = 0 : N-1$) that indicate that $\boldsymbol{p}_k$ are in the polyhedron $p$. We force all the points to be in at least one of the polyhedra with the constraint $\sum_{p=0}^{P_k-1} b_{kp} \geq 1$. 

\subsubsection{Formulation} \label{sect:formulation}
We formulate our MPC under the following Mixed-Integer Quadratic Program (MIQP) formulation. We remove the superscript $l$ (which indicates the number of the iteration) from the reference and state variables for simplification.
 
\begin{align}
    & \underset{\substack{\boldsymbol{x}_k,\boldsymbol{u}_k}}{\text{minimize}}
& & \sum_{k=0}^{N-1} (||\boldsymbol{x}_k - \boldsymbol{x}_{k,ref}||_{\boldsymbol{R}_x}^2 +  ||\boldsymbol{u}_k||_{\boldsymbol{R}_u}^2) \nonumber \\
& & & + ||\boldsymbol{x}_N - \boldsymbol{x}_{N,ref}||_{\boldsymbol{R}_N}^2 \label{eq:NLP}\\
& \text{subject to} & & \boldsymbol{x}_{k+1} = f_d(\boldsymbol{x}_k,\boldsymbol{u}_k),\quad k = 0:N-1 \label{eq:k}\\
& & & \boldsymbol{x}_0 = \boldsymbol{X}_0  \\
& & & \boldsymbol{v}_N = \boldsymbol{0}  \\
& & & |a_{x,k}| \leq a_{x,max} \\
& & & |a_{y,k}| \leq a_{y,max}, \quad a_{z,k} \leq a_{z,max} \\
& & & a_{z,k} \geq a_{z,min}, \quad |j_{x,k}| \leq j_{x,max} \\
& & & |j_{y,k}| \leq j_{y,min}, \quad |j_{z,k}| \leq j_{z,max} \\
& & & b_{kp} = 1 \implies \boldsymbol{A}_{kp}\boldsymbol{p}_k \leq \boldsymbol{c}_{kp} \label{eqn:const_poly} \\
& & & \sum_{p=0}^{P_{k}-1} b_{kp} \geq 1 \\
& & & b_{kp} \in \{0,1\}
\end{align}

The reference trajectory $\boldsymbol{x}_{k,ref}$ is generated as described in Sect. \ref{sect:ref_path}. $\boldsymbol{R}_x$, $\boldsymbol{R}_N$ and $\boldsymbol{R}_u$ are the weight matrix for the discrete state errors without the final state, the weight matrix for the final discrete state error (terminal state), and the weight matrix for the input, respectively.

This optimization problem is solved at every planning iteration to generate an optimal trajectory with respect to its cost function. The MIQP is solved using the Gurobi solver \cite{gurobi}.

\begin{table*}[ht]
\centering
\begin{tabular}{c | cccccc}\hline
 & Success & Flight distance (m) & Flight velocity (m/s) & Flight time (s) & Computation time (ms) & Jerk cost (10\textsuperscript{3}m/s\textsuperscript{3})\\ \hline
Our planner & 5/5 & 51 / 53.83 / 1.8 & 2.75 / 4.37 / 1.32 & 18.5 / 19.8 / 0.9 & 37 / 93 / 13 & 1.01 / 1.63 / 0.45 \\ \hline
\end{tabular}

\caption{Results on 5 randomly generated maps of size $50 \ m \times 12\ m\times 12\ m$ and with 250 dynamic obstacles/250 static obstacles. We show the \textbf{mean / max / standard deviation} of each metric.}
\label{tab:res}
\end{table*}

\section{Limitations}
Our method constrains each point to be in a polyhedron, unlike \cite{toumieh2020planning}, \cite{toumieh2021multiagent} which constrain the whole segment to be in a polyhedron. We opted for this choice because during our testing, the solver was not finding solutions/converging in some cases when we constrained the whole segment to be in a polyhedra instead of just the point. This results in our method requiring the obstacles be further inflated to avoid collision when the segment between 2 points in 2 different polyhedra passes through an obstacle.

Furthermore, the method we presented is based on MPC and this implies that the length of the horizon will affect its performance: the method may fail to find a solution when the dynamics of the agent do not allow it to avoid the moving obstacles using only the chosen time horizon. Increasing the time horizon would increase trajectory quality and reduce the amount of times it fails to find a solution, but this comes at a heavy computational cost.

\section{Simulation}
\subsection{Simulation environment}
The simulation is done in a $50\ m\times12\ m\times12\ m$ environment that is represented in a voxel grid \cite{toumieh2020mapping} of voxel size $0.3$. It contains 200 static obstacles and 200 dynamic obstacles that can fit inside a cube of side length $1.5 \ m$ i.e. 5 voxels. Each voxel inside the obstacle cube is occupied with a probability of 0.1. Then the obstacles are inflated by a voxel to account for the drone radius. The dynamic obstacles oscillate at a random frequency between $\pi/4 \ rad/s$ and $\pi/7 \ rad/s$ along a line whose position, direction and length (between $0 \ m$ and $5 \ m$) are generated randomly (following a uniform distribution). The Gurobi solver is set to use one thread
only as this resulted in faster computation times during our simulations.
All testing is done on the Intel Core i7-9750H up to 4.50 GHz CPU.

\subsection{Planner parameters}
We choose the following parameters: $N = 7$, $h = 100 \ ms$, $g = 9.81 \ m/s^2$, $a_{x,max} = a_{y,max} = 2*g$, $a_{z,max} = g$, $a_{z,min} = -g$, $j_{x,max} = j_{y,max} = j_{z,max} = 90 \ m/s^2$, $v_{max} = 4 \ m/s$, $\boldsymbol{D}_{lin\_max} = diag(1,1,1)$, $thresh\_dist = 0.4 \ m$. The weight matrices are diagonal: $\boldsymbol{R}_x = diag(5,5,5,0,0,0,0,0,0)$, $\boldsymbol{R}_N = diag(50,50,50,0,0,0,0,0,0)$ and $\boldsymbol{R}_u = diag(0.005,0.005,0.005)$. 
The drone radius is $d_{rad} = 0.2 \ m$.

The DMP planner pushes the JPS path $0.4\ m$ away from obstacles (when possible). The path finding step is run only once at the beginning of the planning since we know the environment beforehand, which means the path finding step has no contribution to the total computation time. The planning time horizon of our planner ($N*h = 0.7 \ s$) cannot be too long because of computation time, nor too short because the agent has to stop at the end of the trajectory and so if it is too short, the velocity will be low. We find a compromise experimentally.

\subsection{Simulation results}
The agent goes from the start position of $(1,6,6)$ to $(49,6,6)$ while avoiding both static and dynamic obstacles. We show in Tab. \ref{tab:res} the results of 5 randomly generated environments. The agent succeeds in avoiding all obstacles and reaching the goal in all 5 simulations. The mean flight distance is $51 \ m$, the mean flight velocity is $2.75 \ m/s$ and the mean flight time is $18.5 \ s$. The mean computation time is $37 \ s$ and the max is $93\ s$. This means that even in the worst case, the computation time is smaller than the time step of the MPC. In case the computation time is higher at a given iteration, we discard the generated solution and plan the next iteration starting from the second position of the trajectory i.e. we assume the robot continued on its trajectory for a 2 steps without replanning.

\begin{figure}
\centering
\includegraphics[trim={0cm 0cm 0cm 0cm},clip,width=0.9\linewidth]{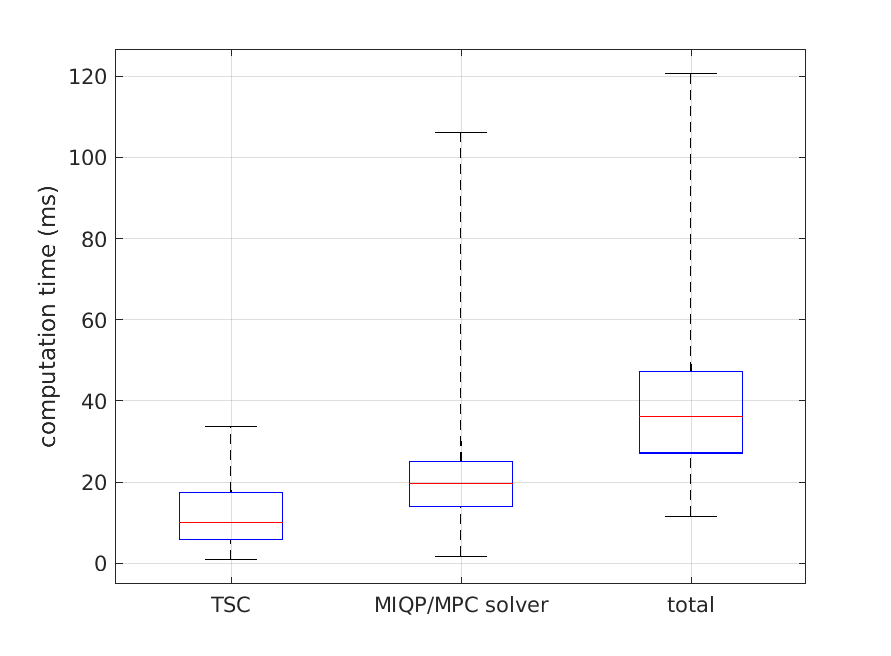}
\caption{Box plots of the computation time for the Safe Corridor generation step, the solver optimization step and the total computation time which includes both steps. The red segment represents the median. The lower and upper bounds of the box represent the 25\textsuperscript{th} and 75\textsuperscript{th} percentile respectively, and the lower and upper whiskers represent the minimum and maximum respectively.}
\label{fig:comp_time}
\end{figure}

We also run another simulation with only dynamic obstacles (for better visibility and performance assessment) with the same parameters as before but we change their positions to be at the same height as the agent and the oscillating direction to be along the $y$ axis. The agent still has the same starting position but the goal is now $(99,6,6)$ (the environment has doubled in size in the $x$ direction). We also change $N = 8$ which increases the time horizon, and consequentially the computation time. The computation times for the Temporal Safe Corridor generation, the MIQP/MPC solver and the total computation time are shown in Fig. \ref{fig:comp_time}. In this case there were a few instances where the computation time exceeded the time step. The performance of the agent is shown in the video \url{https://youtu.be/2ha8Huqi_qI}.

\section{Conclusions and Future Works}
In this paper, we explored a novel planning methods for multirotors in dynamic environments. The method consists in generating a Temporal Safe Corridor (which we defined in this paper), a local reference trajectory, and finally using the TSC and the reference trajectory in an MIQP/MPC formulation to generate a safe trajectory that avoids all types of obstacles (static and dynamic). The method is tested in simulations and the limitations and challenges are presented.

In the future, we plan on investigating a more efficient way to partition the free space around a robot and generate a temporal safe corridor with a minimal number of polyhedra to have a faster optimization time/computation time.
This would require an approach similar to \cite{toumieh2020planning}. At every planning iteration, we would have to update a local voxel grid with the latest measurements from the sensors, generate a global path and a safe corridor (stage 1 now runs every iteration), and then proceed with stage 2.






\bibliographystyle{IEEEtran}
\bibliography{IEEEabrv,IEEEexample}

\end{document}